\documentclass[10pt,twocolumn,letterpaper]{article}

\usepackage[]{wacv}              

\usepackage{graphicx}
\usepackage{subcaption} 
\usepackage{amsmath}
\usepackage{amssymb}
\usepackage{booktabs}
\usepackage{makecell}
\usepackage{multicol}
\usepackage{xcolor}
\usepackage{comment}
\usepackage{float}

\usepackage[pagebackref,breaklinks,colorlinks]{hyperref}

\usepackage[capitalize]{cleveref}
\crefname{section}{Sec.}{Secs.}
\Crefname{section}{Section}{Sections}
\Crefname{table}{Table}{Tables}
\crefname{table}{Tab.}{Tabs.}

\def\wacvPaperID{1378} 
\def\confName{WACV}
\def\confYear{2026}
\def\systemname{FlyPose}

\begin{document}

\title{FlyPose: Towards Robust Human Pose Estimation From Aerial Views} 
\author{
Hassaan Farooq \and Marvin Brenner \and Peter Stütz \and  \\
Universität der Bundeswehr Munich \\
{\tt\small \{hassaan.farooq, marvin.brenner, peter.stuetz\}@unibw.de}
}
\maketitle

\begin{abstract}

Unmanned Aerial Vehicles (UAVs) are increasingly deployed in close proximity to humans for applications such as parcel delivery, traffic monitoring, disaster response and infrastructure inspections. Ensuring safe and reliable operation in these human-populated environments demands accurate perception of human poses and actions from an aerial viewpoint.  This perspective challenges existing methods with low resolution, steep viewing angles and (self-) occlusion, especially if the application demands realtime feasible models. We train and deploy \textit{FlyPose}, a lightweight top-down human pose estimation pipeline for aerial imagery. Through multi-dataset training, we achieve an average improvement of 6.8 mAP in person detection across the test-sets of Manipal-UAV, VisDrone, HIT-UAV as well as our custom dataset. For 2D human pose estimation we report an improvement of 16.3 mAP on the challenging UAV-Human dataset. FlyPose runs with an inference latency of {\raise.17ex\hbox{$\scriptstyle\sim$}}20 milliseconds including preprocessing on a Jetson Orin AGX Developer Kit and is deployed onboard a quadrotor UAV during flight experiments. 
We also publish \textit{FlyPose-104}, a small but challenging aerial human pose estimation dataset, that includes manual annotations from difficult aerial perspectives:
\href{https://github.com/farooqhassaan/FlyPose}{\color{blue}https://github.com/farooqhassaan/FlyPose}.


\end{abstract}
    
\section{Introduction}
\label{sec:intro}

\begin{figure}[tbp] 
  \centering
  \vspace{30pt}
  \begin{subfigure}[b]{0.235\textwidth}
    \centering
    \includegraphics[width=\linewidth]{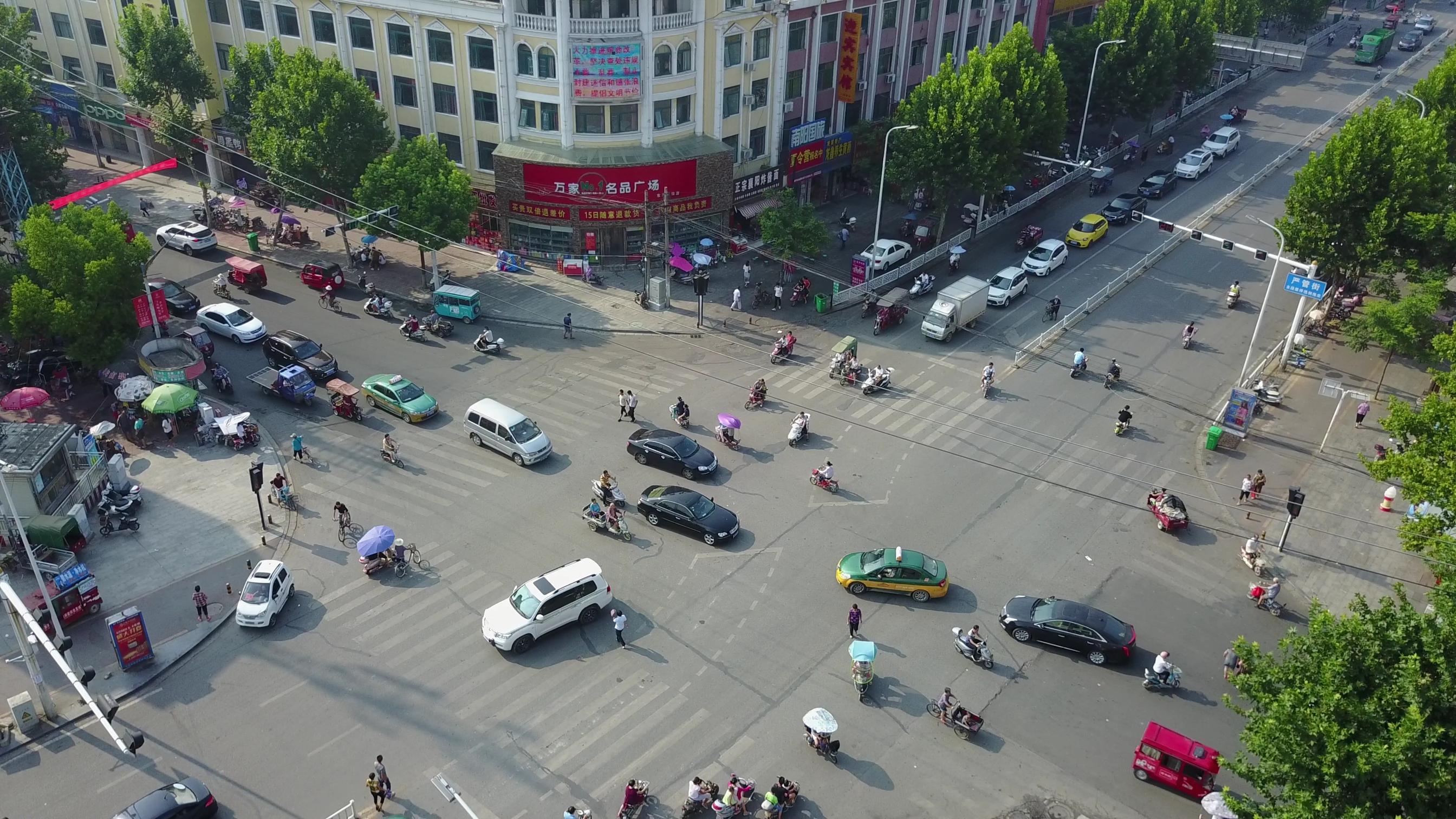}
  \end{subfigure}
  \hfill
  \begin{subfigure}[b]{0.235\textwidth}
    \centering
    \includegraphics[width=\linewidth]{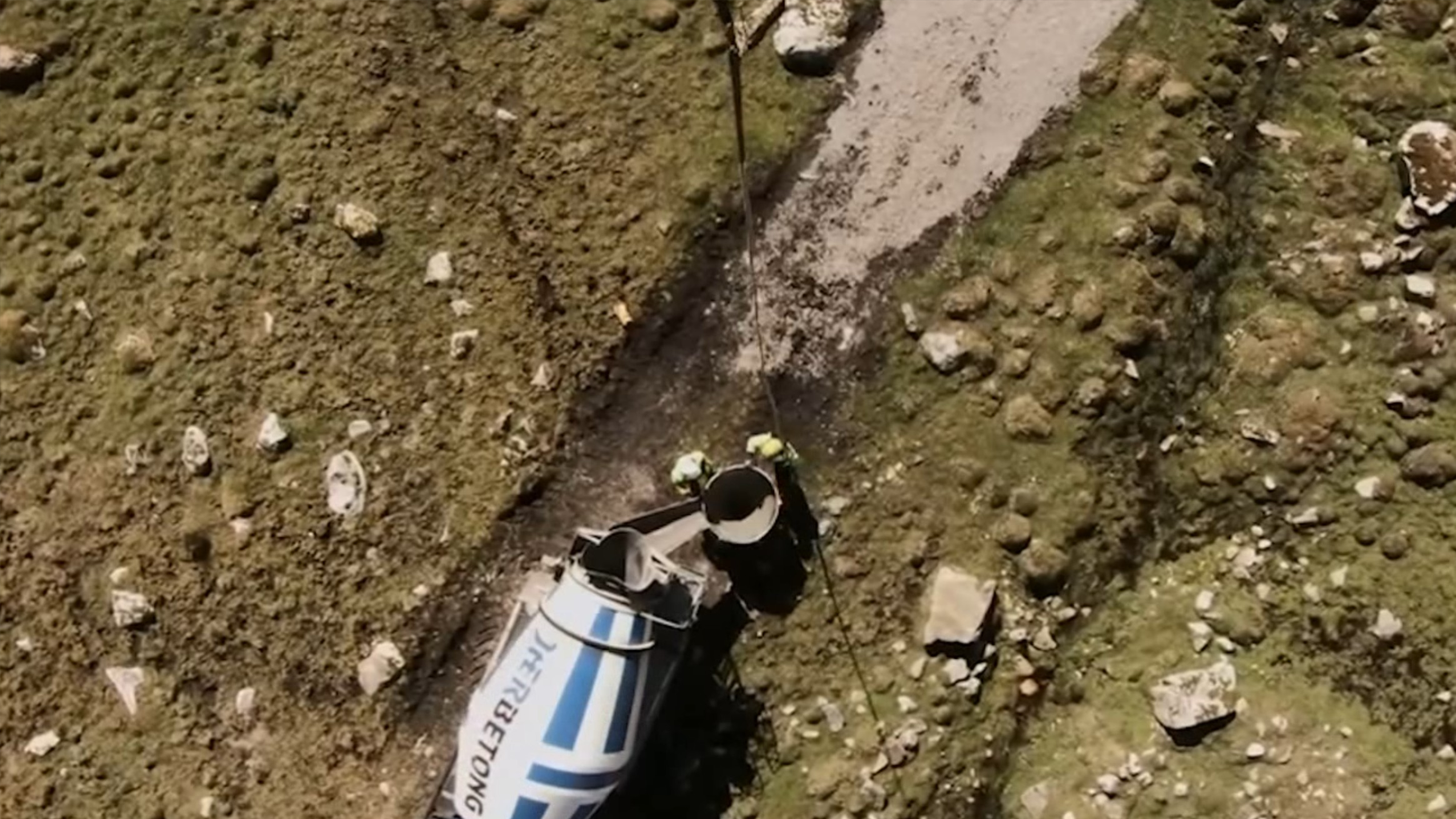}
  \end{subfigure}
  \hfill
  \begin{subfigure}[b]{0.235\textwidth}
    \centering
    \includegraphics[width=\linewidth]{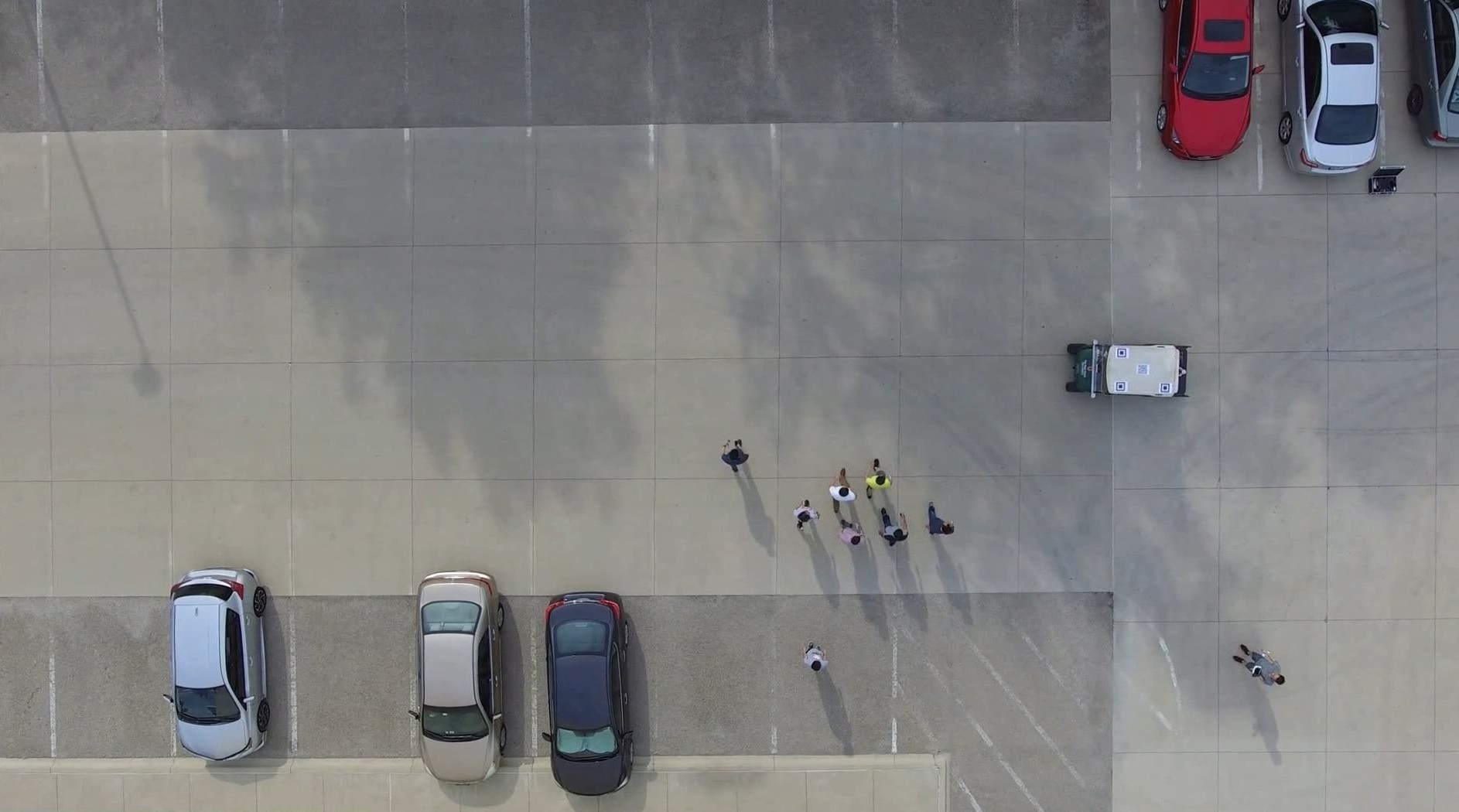}
  \end{subfigure}
  \hfill
  \begin{subfigure}[b]{0.235\textwidth}
    \centering
    \includegraphics[width=\linewidth]{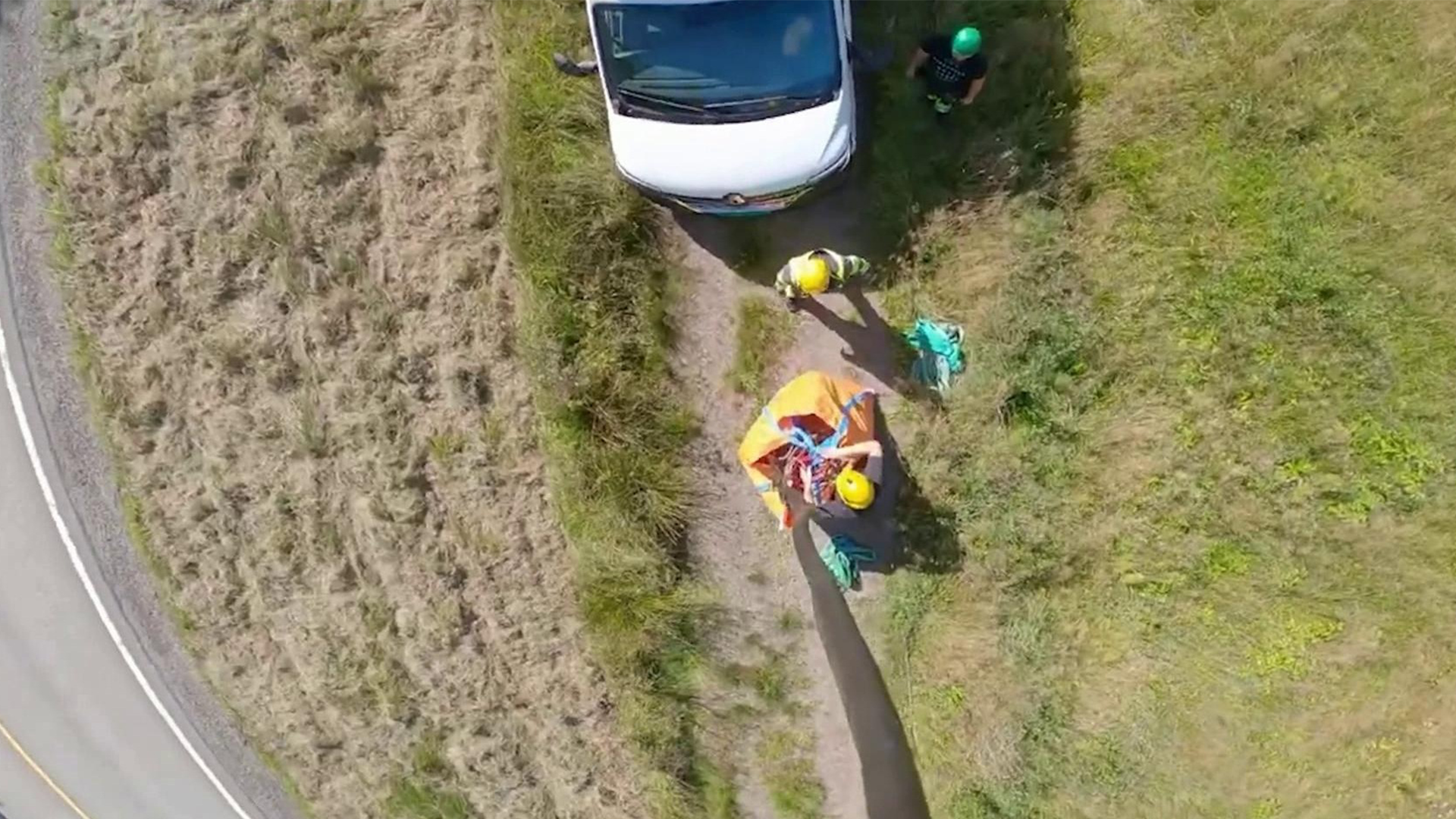}
  \end{subfigure}  
  \centering
  \caption{Applications like urban traffic monitoring and sling-load cargo deliveries with unmanned aerial vehicles pose a major challenge for human pose estimation due to varying person sizes, occlusion and low-resolution overhead imagery \cite{brenner2023udw,du2019visdrone, yang2019top}.}
  \label{fig:intro_scenarios}
\end{figure}

With rising levels of automation in commercial unmanned aerial vehicles (UAVs, also commonly referred to as drones), new applications such as parcel delivery, search-and-rescue, infrastructure inspection and urban traffic monitoring are being introduced. Their operation in human-populated spaces \cite{baytas2019design} makes the robust perception of human presence and behaviors indispensable. In both human-human and human-machine interaction, the configuration of body joints conveys non-verbal cues for understanding body language, physical movement and intent. The task of human pose estimation (HPE) represents a widely-used foundational module to extract these cues from images to enable skeleton-based downstream tasks such as action recognition, pose-based tracking, motion prediction, engagement detection or interaction via hand gestures. Based on the insights gained from these tasks, the UAV can then respond appropriately based on its own behavior model. 
The majority of applications for HPE cannot assume a ground-level or shoulder-height camera view where the human subject is seen frontally or laterally, as this viewpoint would enable the observation of distinguishable human body features. 
Aerial viewpoints tend to occlude pose keypoints of the face or legs, distort body proportions and challenge pose models trained on ground-level images. Therefore, our work explores human pose estimation from an aerial perspective to aid UAV applications in human-populated environments (Figure \ref{fig:intro_scenarios}). %

\begin{figure}[hbt]
  \centering
  \begin{subfigure}[b]{0.2\textwidth}
    \centering
    \includegraphics[width=\linewidth]{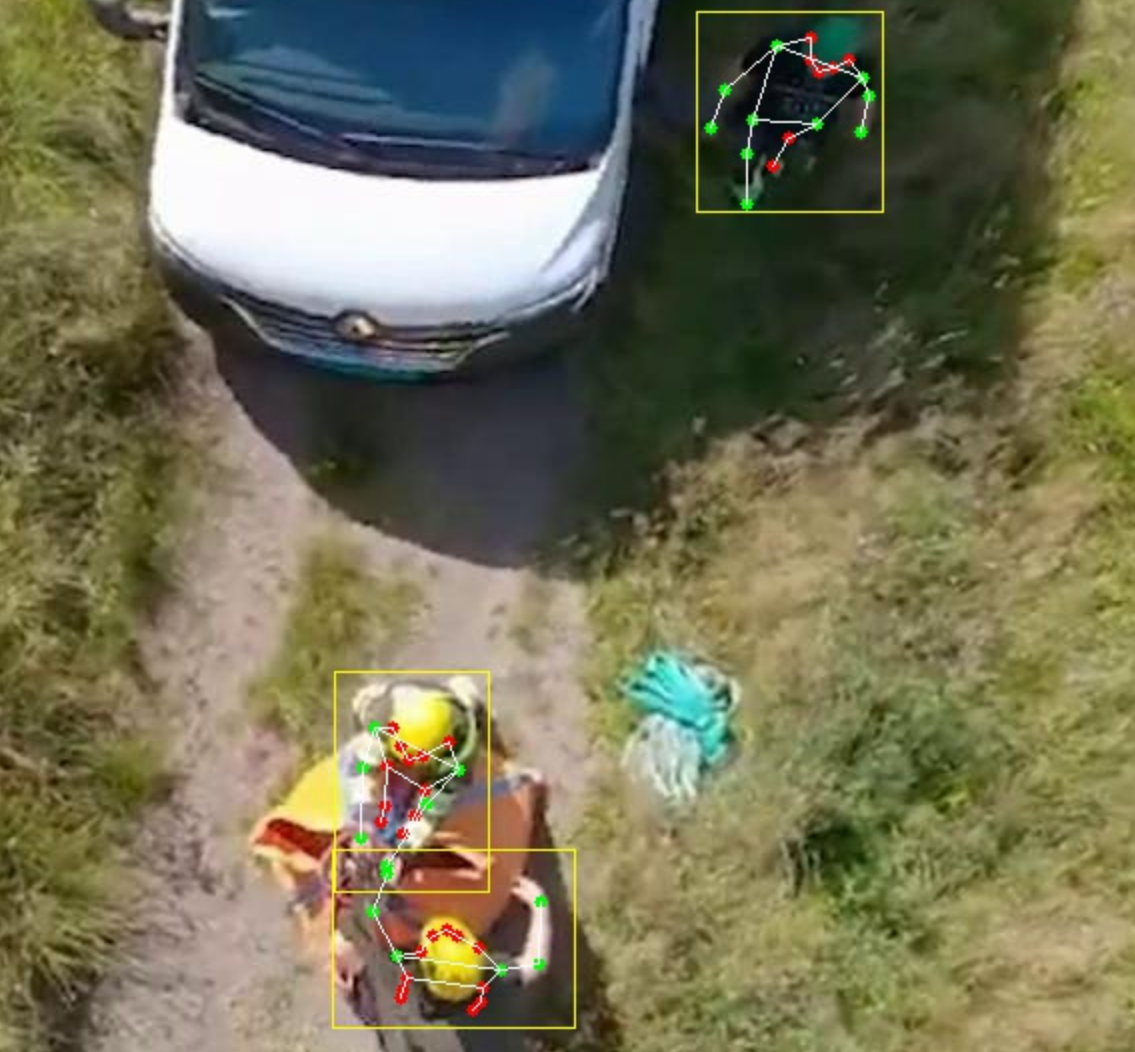}
  \end{subfigure}
  \hfill
  \begin{subfigure}[b]{0.25\textwidth}
    \centering
    \includegraphics[width=\linewidth]{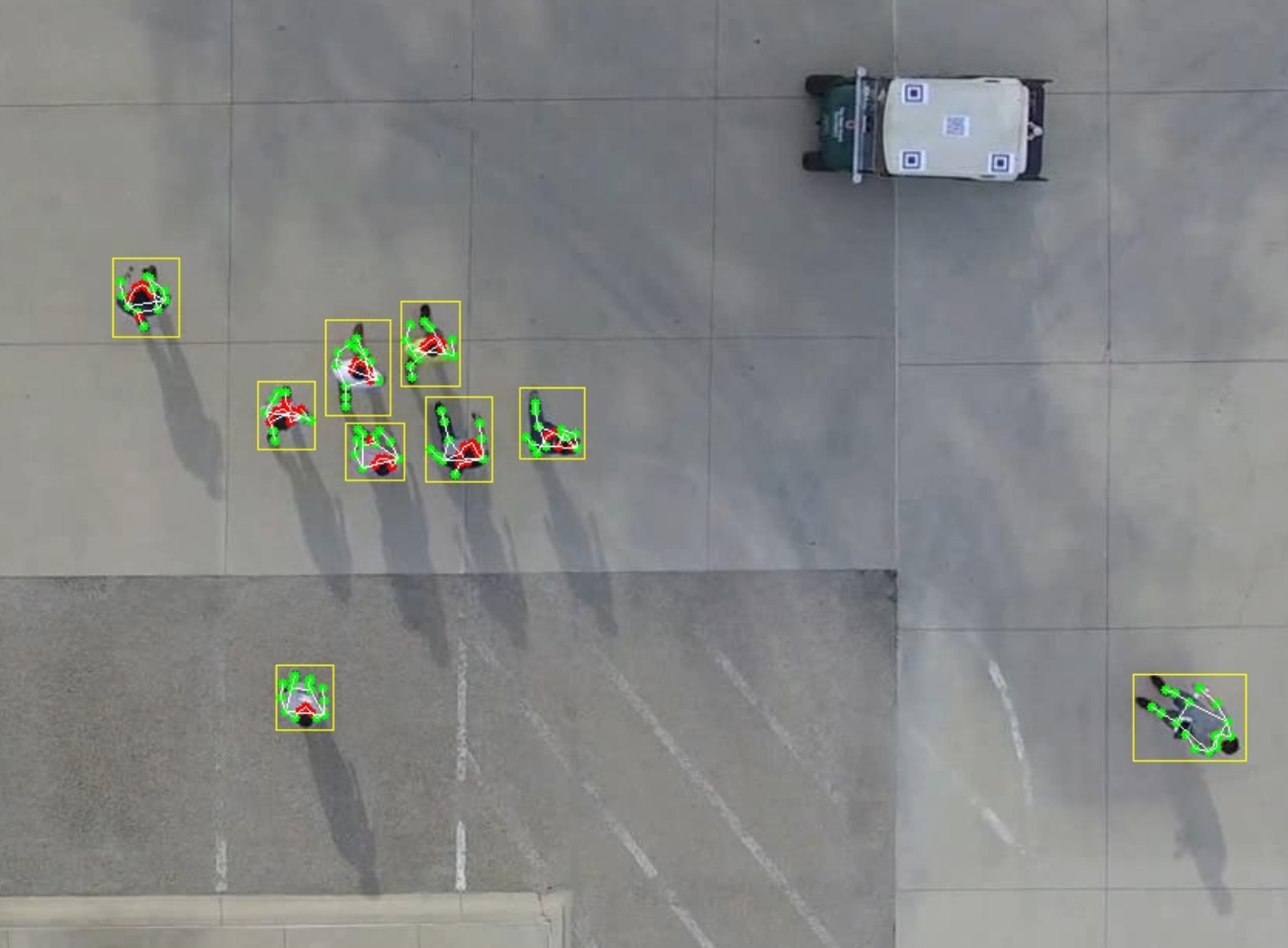}
  \end{subfigure}
  \caption{Two examples from our FlyPose-104 dataset with manually annotated bounding boxes and poses, featuring frequent self-occlusions of lower body and facial joints (marked in red) \cite{yang2019top,brenner2023udw}. }
  \label{fig:qualitative_motivation}
\end{figure}

Aerial human pose estimation introduces distinct challenges compared to ground-based setups. The UAV's sensor depression angle is steep, up to 90 degrees, resulting in top-down imagery of humans with foreshortened limbs and frequent self-occlusion (Figure \ref{fig:qualitative_motivation}). At the same time, the UAV needs to ascend to altitudes free of obstacles, where its sensor's ground sampling distance increases and hence people occupy less pixels in the image. 
This drastically limits the distance at which the sensor can perceive humans and the model is able to predict stable, usable poses. Additionally, UAVs must operate under strict payload weight, dimensions and power constraints during flight, further limiting available computational resources for processing higher resolution images onboard. \\

In this paper, we investigate these challenges by training \systemname, a top-down human pose estimation model with a focus on aerial imagery that is deployable on edge-devices. Our contributions are as follows: 

{
\begin{itemize}
	\item Development of a top-down human pose estimation pipeline to predict more accurate poses from aerial views by training across multiple aerial datasets to deal with top-down views and small scales.
	\item Manual annotation and release of the FlyPose-104 dataset, a challenging aerial pose estimation test set.  
	\item Integration of the trained models into a hardware system, with reporting of model latencies on an edge device during real flight.
\end{itemize}}

\section{Related Work}
\label{sec:relatedwork}

Top-down pose estimation methods like \cite{xu2022vitpose, alphapose} begin by detecting person bounding boxes using an object detector, and then run a body keypoint estimation model on each cropped region. This allows for higher precision compared to bottom-up methods like \cite{cao2019openposerealtimemultiperson2d,lu2024rtmo}, as the pose model focuses on a smaller, person-specific patch with less distractions from unconstrained backgrounds.
Near ground-level, VisionTransformer \cite{visiontransformer} based object detectors like Co-DETR \cite{zong2023detrs} can already localize humans robustly across a variety of scene backgrounds \cite{lin2014microsoft}. 

For multi-scale aerial imagery in altitudes above 20 meters, low resolution, dense object clusters and occlusion impede large scale data annotation, as a consequence the publicly available training data is typically scattered into smaller application-specific datasets, like urban surveillance \cite{du2019visdrone, liu2021visdrone}, traffic monitoring  \cite{yu2020unmanned}, Search and Rescue \cite{bovzic2019deep, zhangRobustAerialPerson2024, varga2022seadronessee}, Helicopter Cargo Operations \cite{brenner2023udw}  and Remote Sensing \cite{wang2021tiny,xia2018dota,lam2018xview}. Tiny object detection requires specifically tailored two-stage architectures \cite{qiao2021detectors, ren2016faster,xu2022detecting} which limit realtime feasibility.
Since extracting human poses is not feasible for tiny person scales, we focus on popular one-stage object detectors like YOLO \cite{redmon2016you} and the few realtime feasible DETR variants \cite{chen2024lw, zhao2024detrs} so far. They are frequently modified to deal with aerial imagery \cite{chen2023yolo}. There have also been attempts to use augmentation strategies like Unified Foreground Packaging \cite{huang2022ufpmp, liu2024esod}, to improve multi-scale loss convergence \cite{singh2018sniper} and reduce the amount of negative background pixels on higher resolution images during training.

However, on a dataset like VisDrone \cite{du2019visdrone}, the performance of the state-of-the-art ViT-YOLO \cite{zhang2021vit} at 41 mAP on the VisDrone test set highlights limited success from the aerial perspective. 
At present, the higher accuracy for smaller objects requires larger input sizes and model complexity, which in turn increases the inference latency. Similar to autonomous driving \cite{zhang2024pedestrian}, we envision that a more generalizable, lightweight person detector would be beneficial for 
practitioners and could be used across UAV applications.

For Pose Estimation, ViT-based architectures like ViTPose \cite{xu2022vitpose} also represent the current state-of-the art on the COCO test-dev set, while examples for lightweight architectures with competitive performance are RTMO \cite{lu2024rtmo}, RTMPose \cite{rtmpose}, KAPAO \cite{mcnally2022rethinking} and YOLO-Pose \cite{maji2022yolo}. Nevertheless, their accuracy drops notably when applied to UAV imagery, due to viewpoint shifts and scale variation, even when not accounting for small objects. Top-down views like in CMU Panoptic \cite{panopticdataset}, ITOPS \cite{haque2016towards, garau2021deca} and PoseFES \cite{yu2023human} are rarely explored outside controlled conditions. There have been very few recent works \cite{Kalampokas2023uavbenchmarking,hwang2024wacvaerial3Dhpe} that try to address the challenge of aerial human pose estimation from UAVs. In \cite{Kalampokas2023uavbenchmarking}, they try to evaluate the performance of out-of-the-box pose estimators on various datasets and gauge their performance on edge devices. The authors of \cite{hwang2024wacvaerial3Dhpe} instead attempt to refine an estimated pose to resemble poses from an aerial poses' latent codebook. 
A small number of aerial datasets like Aerial Gait \cite{aerialgaitdataset} or the synthetic dataset in Airpose \cite{airposedataset} exist, but they are limited in variety and scale. Airpose attempts multi-view 3D human pose and shape estimation but the used methods still rely on 2D pose priors from OpenPose and AlphaPose which were only finetuned for slightly elevated views from the side without top-views or covering larger distances. UAV-Human \cite{li2021uav} is the largest aerial pose dataset so far, offering annotated human keypoints from drone footage in diverse settings. 

The models and datasets in the literature have potential for improvement, as due to viewpoint shifts and scale variation, the detection and 2D poses often fail on real-world UAV topview imagery like \cite{russell2024nomad, brenner2023udw}. This underscores the need for a HPE model designed for aerial perspectives. Our work seeks to bridge this gap by adapting and extending high-performing top-down models to perform robustly in diverse UAV-captured environments under realtime constraints. We also deploy our pipeline in real flight onboard a UAV to validate the results.

\section{Methodology}
\label{sec:method}


%

\begin{figure}[htbp]
  \centering
  \includegraphics[width=\linewidth]{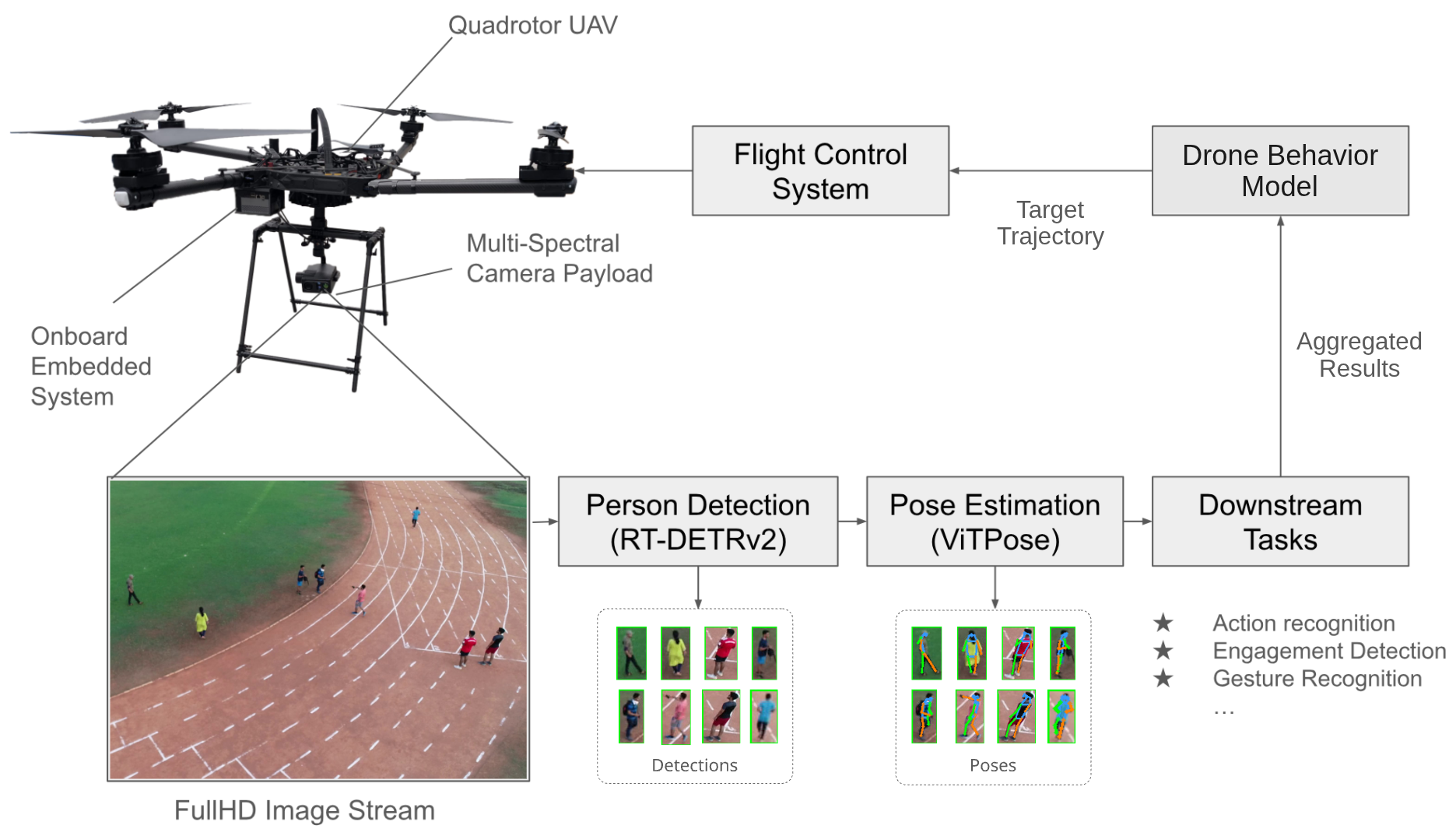}
   \caption{System overview: the bottom illustrates the FlyPose pipeline, where the detector and pose estimation model are trained separately, and the top is an example of how the aggregated information can be used for downstream tasks within the drone system for various applications.}
   \label{fig:system_diagram}
\end{figure}

We present \textit{FlyPose}, an onboard human pose estimation approach tailored for drones operating under tight computational constraints in altitudes of up to 40 meters. FlyPose follows a top-down human pose estimation (HPE) pipeline where incoming Full-HD video frames from a stabilized, multi-spectral camera are first processed by a lightweight person detector, after which a dedicated pose estimation module extracts 2D keypoints from the detected regions. An overview of the system is illustrated in Figure~\ref{fig:system_diagram}.\\

FlyPose is designed to serve as a perceptual backbone for a range of person-aware downstream tasks such as gesture-based control, action recognition and future pose prediction. The insights from the downstream tasks are intended to be integrated into a broader drone behaviour model, which can then instruct the flight controller to move along a desired trajectory or alter the current maneuver.
In this work, we focus on the core challenge of achieving reliable onboard 2D human pose estimation in realtime. We choose a lightweight top-down HPE approach, where the detector is separately trained for aerial person detection and a heatmap-based pose estimation method is finetuned to deal with unusual, self-occluding overhead poses. We leverage a combination of existing datasets to achieve better performance on small scale persons and overhead views. Generally, our strategy is to first benchmark model architectures on aerial datasets to identify suitable models regarding the accuracy-latency trade-off and then finetune on additional aerial data to improve the results for our application without introducing significant computational overhead. 
After FlyPose is trained, we optimize the model for deployment on an edge device and test it in real flight experiments.

\subsection{Top-down Human Pose Estimation}
The person detector receives an input image \( I \in \mathbb{R}^{H \times W \times 3} \) and outputs a set of bounding boxes \(B\), where each bounding box \( b_i = (x_i, y_i, w_i, h_i) \) represents the coordinates of the top-left corner and the width and height of the detected person. We follow the training procedure of RT-DETRv2 \cite{lv2024rt}, but experiment with replacing the generalized intersection over union loss with Normalized Wasserstein Distance Loss (NWDL) \cite{xu2022detecting} to achieve more stable training despite the smaller objects in aerial imagery, similar to \cite{brenner2023udw}. 

The pose estimation is then performed on cropped image patches corresponding to each person detection \(b_i\). The pose estimation task is framed as the prediction of heatmaps for the 17 body keypoints defined by the COCO format. For each keypoint, the network outputs a 2D confidence map, indicating the likelihood of the keypoint's presence at each pixel location. The training objective minimizes the discrepancy between the predicted and ground-truth heatmaps using the standard mean squared error (MSE) loss. During inference, keypoints are extracted from these heatmaps using non-maximum suppression (NMS) to localize peak responses. 


\subsection{Evaluation Metrics}
For person detection, we use the COCO mean Average Precision (mAP) which averages the Average Precisions across the 0.5-0.95 intersection over union (IoU) thresholds and the Average Recall (AR) for 100 allowed predictions. Since we conduct multi-dataset training, the Average Precision and Recall across all testing sets, weighted by the number of corresponding frames, are reported for reference. We do not report classification scores, since all datasets were filtered to only include one person class. For human pose estimation, we evaluate the model performance using standard COCO mean Average Precision (mAP) for keypoints, similarly averaging across the 0.5-0.95 Object Keypoint Similarity (OKS) thresholds. We use the standard OKS metric by COCO \cite{lin2014microsoft} that measures how close predicted keypoints are to the ground-truth, normalized by the person scale (bounding box area) and a per-keypoint tolerance.
To measure the feasibility of the models in practice, we measure the latency of the models as the time taken for a single forward pass through the model with a batch size of 1. The latency of the person detection and pose estimation models are reported separately, with the expectation of the latency of pose estimation to increase for multiple persons because of the top-down pose estimation. Although this can be offset by batching person detections together and incurring a cost of preparing the batch, further experimentation is necessary. The inference latency for the models is measured on a Mobile GPU A6000 as well as on the Jetson Orin AGX Developer Kit with 32 GB RAM. Note that on the Jetson, the models were converted to FP32 TensorRT engines for lower latency. 

\subsection{FlyPose-104 dataset}
\label{subsection_flypose104}
Due to the limited availability of annotated human pose data for aerial imagery, we collect and annotate a small dataset of 104 images, including our own images and those sourced from \cite{brenner2023udw, yt_mountains, yt_seadrone, yang2019top}. A few annotated samples of our FlyPose-104 dataset can be seen in Figure \ref{fig:qualitative_motivation}.
It covers a variety of backgrounds including snow, dirt, concrete, water and grass, making it a small yet challenging dataset. The viewpoints feature 90 degrees top-down viewpoints as well as heavy occlusion examples. A total of 193 persons were manually annotated with person bounding boxes and 17 COCO pose keypoints, along with the keypoint visibility flag. We include samples with multi-scale person bounding boxes recorded at altitudes between 5 and 50 meters. 
FlyPose-104 essentially aims to capture representative images that reflect the wide range of challenges inherent to aerial human pose estimation, including variations in person scale (due to altitude), camera viewpoints (top-down and angled), background complexity, application contexts and (self-)occlusions.

\subsection{Preliminary Testing}
\label{subsec:prelim_testing}
For the person detector, RT-DETRv2-S was tested with 
alternative pretrained backbones like RegNetX-400MF \cite{radosavovic2020designing}, Hiera-Tiny \cite{ryali2023hiera} and a bottom-heavy version \cite{ning2023rethinking} of ResNet-18 \cite{he2016deep} by finetuning for 50 epochs on VisDrone2019-DET at 1280 pixels input resolution. However, none of them performed higher on the VisDrone2019-DET test-dev set after training than regular ResNet-18 with 28.6 mAP which is in line with a recent backbone benchmark \cite{goldblum2023battle}. It showed that for generic object detection on limited computational resources, ResNet-18 so far remains an efficient choice preferable to transformer-based backbones.
 When observing qualitative results on VisDrone, we notice that most of the precision loss for tiny backbones comes from smaller objects in the background that they cannot accurately localize, at least in a single stage without bidirectional feature fusion. 
 For RT-DETRv2-S , the preliminary detection mAP values for small (18.4 mAP), medium (40.0 mAP) and large (53.6 mAP) objects on the VisDrone test set support this hypothesis. Therefore, we decide to use ResNet-18 as a backbone for RT-DETRv2 at the cost of missing tiny people in the background, that would likely not result in stable, meaningful poses for downstream tasks anyhow.

\begin{figure}[htb]
  \centering
  \includegraphics[width=\linewidth]{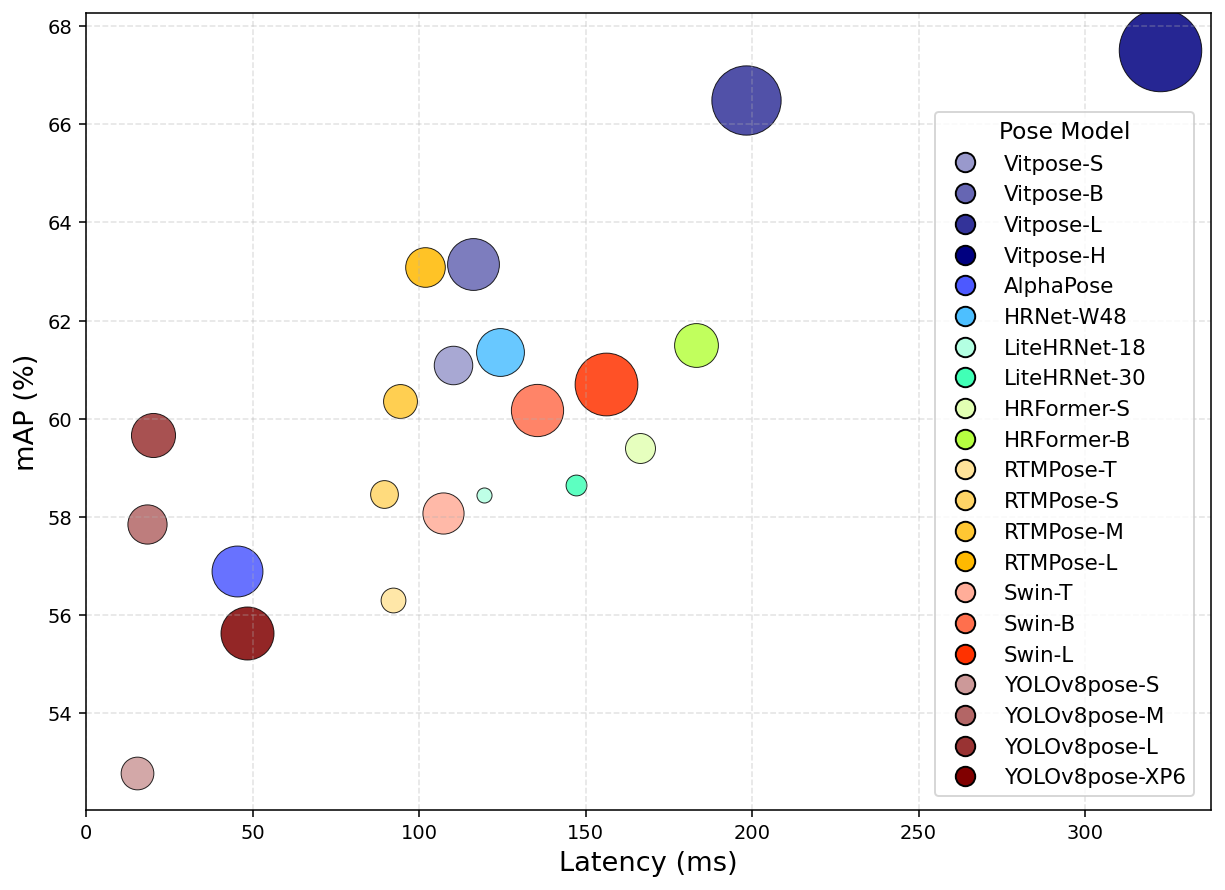}
   \caption{Pose Estimation performance of pretrained models on the UAV-Human dataset, plotted against their latency on an RTX A6000 GPU. Each circle's radius is proportional to the model parameter count.}
   \label{fig:system_latency}
\end{figure}

To evaluate the performance of current state-of-the-art pose estimation models in aerial scenarios, we perform preliminary evaluations on the UAV-Human v1 test set~\cite{li2021uav}. The authors \cite{li2021uav} report results using pretrained AlphaPose and HigherHRNet, with AlphaPose slightly outperforming at 56.9 mAP~\cite{li2021uav}. Building on this, we evaluate a range of recent 2D pose estimation architectures that have shown strong results on ground-level datasets like COCO, including ViTPose \cite{xu2022vitpose}, RTMPose \cite{rtmpose}, AlphaPose \cite{alphapose}, HRNet \cite{SunXLW19}, LiteHRNet \cite{Yulitehrnet21}, Swin-Pose \cite{swinpose} and YOLOv8-Pose \cite{maji2022yolo}. Our aim was to first assess how well these pretrained models perform on the aerial perspectives of the UAV-Human dataset, the results of which can be seen in Figure \ref{fig:system_latency}. Among the tested models, ViTPose-H pretrained on COCO achieved the highest performance with 67.52 mAP on the UAV-Human dataset. RTMPose is a decent competitor for its model size but overall given both ViTPose's strong results in this setting and its leading performance on COCO, we continue with this architecture for our application.
\begin{table*}[t]
  \centering
  {\small{
  \begin{tabular}{l | c | c | c | c | c | c | c | c | c | c | c | c}
   \toprule
   & \multicolumn{2}{c|}{COCO-Person} & \multicolumn{2}{c|}{VisDrone-Person} & \multicolumn{2}{c|}{Manipal-UAV} & \multicolumn{2}{c|}{HIT-UAV-Person} & \multicolumn{2}{c|}{FlyPose-104 } & \multicolumn{2}{c}{Weighted Average} \\
   & \multicolumn{2}{c|}{val} & \multicolumn{2}{c|}{test-dev} & \multicolumn{2}{c|}{test} & \multicolumn{2}{c|}{test (infrared)} & \multicolumn{2}{c|}{test (ours)} & \multicolumn{2}{c}{across test sets} \\
   Method  & AP & AR & AP & AR & AP & AR & AP & AR & AP & AR & AP & AR \\ 
   \midrule
   Baseline & 60.75 & 75.57 & 10.44 & 18.96 & 20.20 & 30.58 & 3.16 & 30.18 & 10.26 & 46.17 & 14.33 & 26.76\\
   + VisDrone   & 21.28 & 43.20 & \textbf{21.08} & \textbf{29.46} & 25.35 & 35.06 & 7.06 & 29.45 & 22.42 & 45.91 & 21.43 & 32.61\\
   + Multi-Dataset & 22.24 & 42.11 & 21.07 & 29.13 & 27.32 & 38.06 & \textbf{52.54} & \textbf{62.73} & 22.67 & 45.13 & \textbf{28.21} & 38.20\\
   + COCO-Person       & \textbf{61.39} & \textbf{75.52} & 20.21 & 28.87 & \textbf{28.39} & \textbf{40.46} & 49.21 & 61.37 & 25.05 & 48.71 & 28.07 & \textbf{39.21}\\
   + NWD Loss     & 60.94 & 74.95 & 20.20 & 28.77 & 27.90 & 40.42 & 49.84 & 61.28 & \textbf{27.41} & \textbf{48.96} & 27.96 & 39.14 \\
   \bottomrule
  \end{tabular}
  }}
  \caption{Person detection performance of RT-DETRv2-S after multi-dataset training and applying NWD loss at 1280 pixels input size. \newline The average across test sets is computed column-wise and weighted by the number of corresponding images to monitor overall performance.}
  \label{tab:example44}
\end{table*}

\section{Experiments and Results}
\label{sec:experiments}

Based on the preliminary testing on the VisDrone and UAV-Human datasets, we continue with the RT-DETRv2 and ViTPose models for the following experiments. We train the aerial person detector and aerial pose estimation networks separately. We evaluate the performance of our trained models, combine the pipeline to test it on edge hardware and finally deploy it in real flight.

\subsection{Aerial Person Detection}
For the quantitative evaluation of person detection, we used the person-only test(-dev) splits of VisDrone \cite{du2019visdrone}, Manipal-UAV \cite{akshatha2023manipal}, HIT-UAV \cite{suo2023hit}
and our custom \textit{FlyPose-104} dataset (Section \ref{subsection_flypose104}).
To monitor the drop in frontal-view generalizability which occurs when finetuning on a larger set of aerial imagery,
we additionally report changes to the COCO-Person validation set performance. The main person detection results after each training are displayed in Table \ref{tab:example44}.

As a first step, RT-DETRv2-S with pretrained weights from COCO \cite{lin2014microsoft} and Objects365 \cite{shao2019objects365} is finetuned for 60 epochs on the VisDrone2019-DET dataset \cite{du2019visdrone} after removing all classes other than ``pedestrian" and ``person", which are combined into a single person class. After the finetuning, the model performs 7.1 mAP higher on average across all of the used test-sets. This first step is standard practice but VisDrone only features imagery from nearby the city of Tianjin, therefore we proceed to gather additional aerial datasets featuring a person class to further improve the model's generalization ability outside urban scenarios. We add the original train/val split of eight additional aerial object detection datasets, resulting in 66849 additional images for training and 21164 for validation besides VisDrone2019-DET. To give a brief overview, we describe them here: SeasDronesSea \cite{varga2022seadronessee} and Heridal \cite{bovzic2019deep} represent RGB aerial imagery for search-and-rescue (SAR) purposes, where the former has maritime and the latter hilly, Mediterranean terrain as background.  
VTSAR \cite{zhangRobustAerialPerson2024}, DroneRGBT \cite{zhang2023drone},  VTUAV-det \cite{zhang2022visible} feature both RGB and thermal imagery from UAVs for training, while HIT-UAV \cite{suo2023hit} is the only pure thermal dataset with a separate testing split. Manipal-UAV \cite{akshatha2023manipal} features person detection at altitudes from 10 to 50 meters with complex backgrounds and TinyPerson \cite{yu2020scale} is the most challenging dataset for our approach, since it focuses purely on dense clusters of tiny people from an aerial perspective. Like with VisDrone, only the person-related classes are kept in the annotations to test if the model learns robust features although presented with both aerial frontal and top-views. Rather than inspecting their individual performance, we combine them into one dataset together with VisDrone and the resulting test set performance is reported in the row ``Multi-Dataset" in Tab. \ref{tab:example44}.  Note that we did not exclude thermal imagery because for nighttime applications like search-and-rescue, the model has to be able to predict on both modalities. As a result, the detector gains another 6.78 mAP from the previous iteration on average but mainly caused by the introduction of the thermal imagery and the resulting improvement in the HIT-UAV test-set. 

\begin{figure}[bp]
  \centering
  \includegraphics[width=\linewidth]{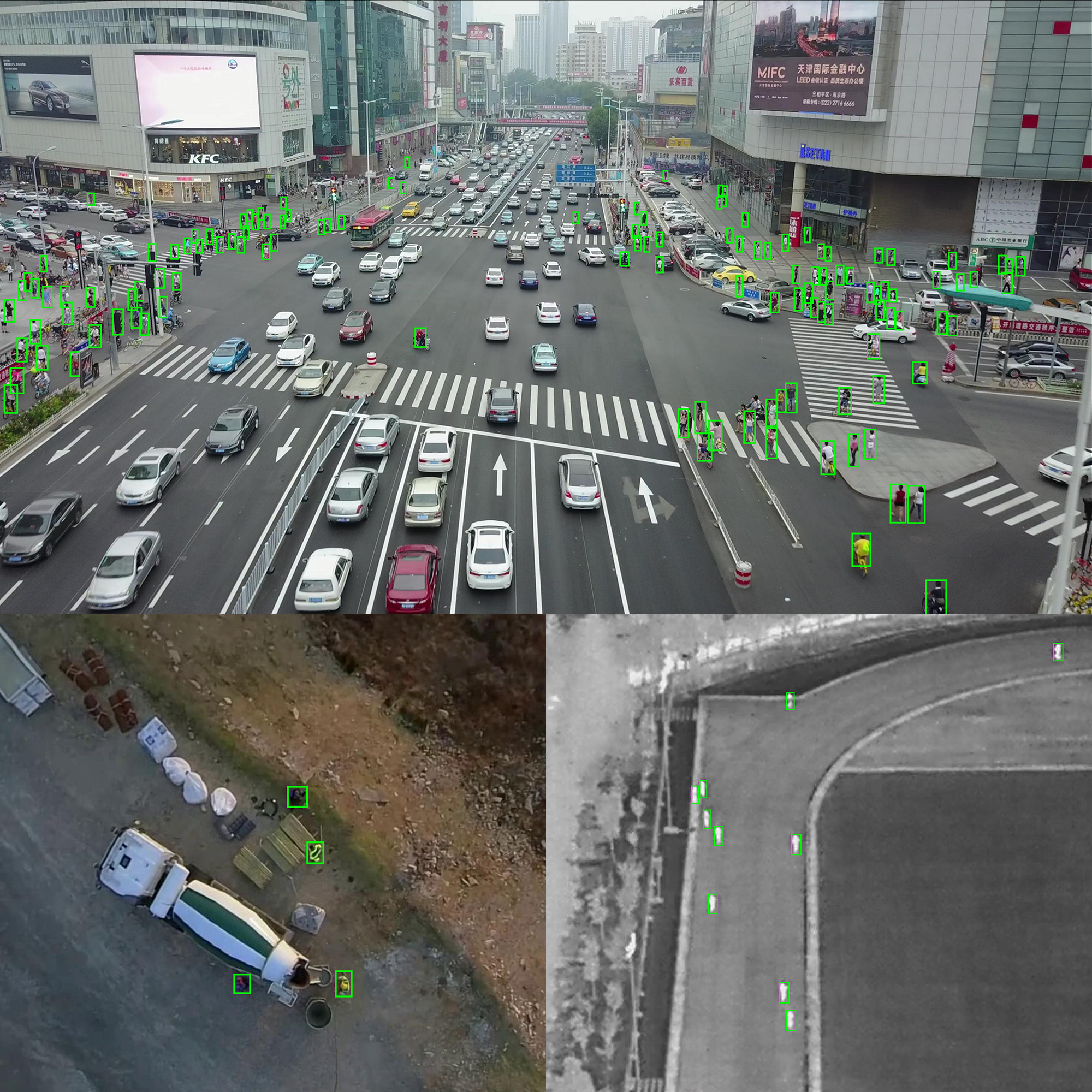}
   \caption{Qualitative detections (red) on the VisDrone2019-DET (top), FlyPose-104 (bottom left) and HIT-UAV (bottom right).} 
   \label{fig:result_detection}
\end{figure}

Since the frontal perspectives within the COCO dataset are still useful in frontal low-altitude perspectives and the initial mAP on the COCO validation set decreased after finetuning on aerial data, we reintroduce the COCO-Person train-val split into the finetuning to mitigate this effect. This step slightly decreased the average mAP on the aerial test-sets by 0.14 mAP but improved the average AR by 1.01. Finally, to improve the localization on smaller objects, Normalized Wasserstein Distance Loss \cite{xu2022detecting} was introduced for the final training which had only a negligible impact overall but it did improve both mAP and AR on our FlyPose-104 test set. Although individual training steps led to the highest increase of mAP on one test-set (for example, only finetuning on VisDrone had the best results on the respective test set), our multi-dataset version of RT-DETRv2-S generalizes robustly across all used aerial test sets (examples in Figure \ref{fig:result_detection}). To establish the model's realtime feasibility, we measured 13 milliseconds
of inference latency for a single forward pass on the Jetson Orin AGX Developer Kit with 32 GB after conversion to a TensorRT FP32 engine.
\subsection{Aerial Pose Estimation}
We experiment with the small, base, large and huge versions of ViTPose, using a consistent input image size of 256x192. During inference, the detected person patch is resized while preserving its aspect ratio with the larger dimension scaled to a fixed size (256px for height or 192px for width), ensuring that the patch is centered without distortion. During training, in addition to the standard augmentation techniques described in ~\cite{xu2022vitpose}, we introduce a downscaling augmentation (by 5-20\% of the patch) to simulate smaller person sizes and lower resolutions encountered in aerial footage during training. As consistent pose annotations for low-resolution people are difficult to obtain, this aims to mimic persons that are viewed from greater distances or affected by motion blur.\\ 
%

\noindent\begin{minipage}{\linewidth}
  \centering
  {\small{
  \begin{tabular}{@{}l c@{} c@{} c@{} c@{}}
    \toprule
    Method& \makecell{mAP\\(pretrained)~~} & \makecell{mAP\\(finetuned)~~} & \makecell{Latency [ms]~~\\(A6000)~~} & \makecell{Latency [ms]~~\\(Jetson Orin)} \\
    \midrule
	ViTPose-S & 61.09 & 65.76 & \textbf{110.23} & \textbf{6.54} \\   
	ViTPose-B & 63.15 & 67.50 & 116.20  & 11.62 \\ 
	ViTPose-L & 66.50 & 70.31 & 198.30 &  22.35 \\
    ViTPose-H & \textbf{67.52} & \textbf{73.18} & 322.55 & - \\
    \bottomrule
  \end{tabular}
  }}
  \captionof{table}{Pose Estimation results of ViTPose on the UAV-Human v1 test-set, using the trained RT-DETRv2-S person detector.}
  \label{tab:example2}
\end{minipage}
\\

We use a similar training strategy to \cite{xu2022vitpose}, starting with COCO-pretrained weights of these models and finetune them for between 170-210 epochs on the UAV-Human v1 train-set. The performance of the pretrained and finetuned models is summarized in Table~\ref{tab:example2}. ViTPose-H achieves the highest accuracy with 73.18 mAP on the UAV-Human dataset which is an improvement of 16.3 mAP over the previously reported AlphaPose result of 56.9 mAP (Section \ref{subsec:prelim_testing}). However, ViTPose-H exhibits significantly slower inference on a single NVIDIA A6000 GPU and we could not deploy it on the Jetson Orin AGX Developer Kit. As expected, using the smaller model sizes leads to lower mAP scores at faster throughput frequencies. 

\begin{figure}[h]
  \centering
  \includegraphics[width=\linewidth]{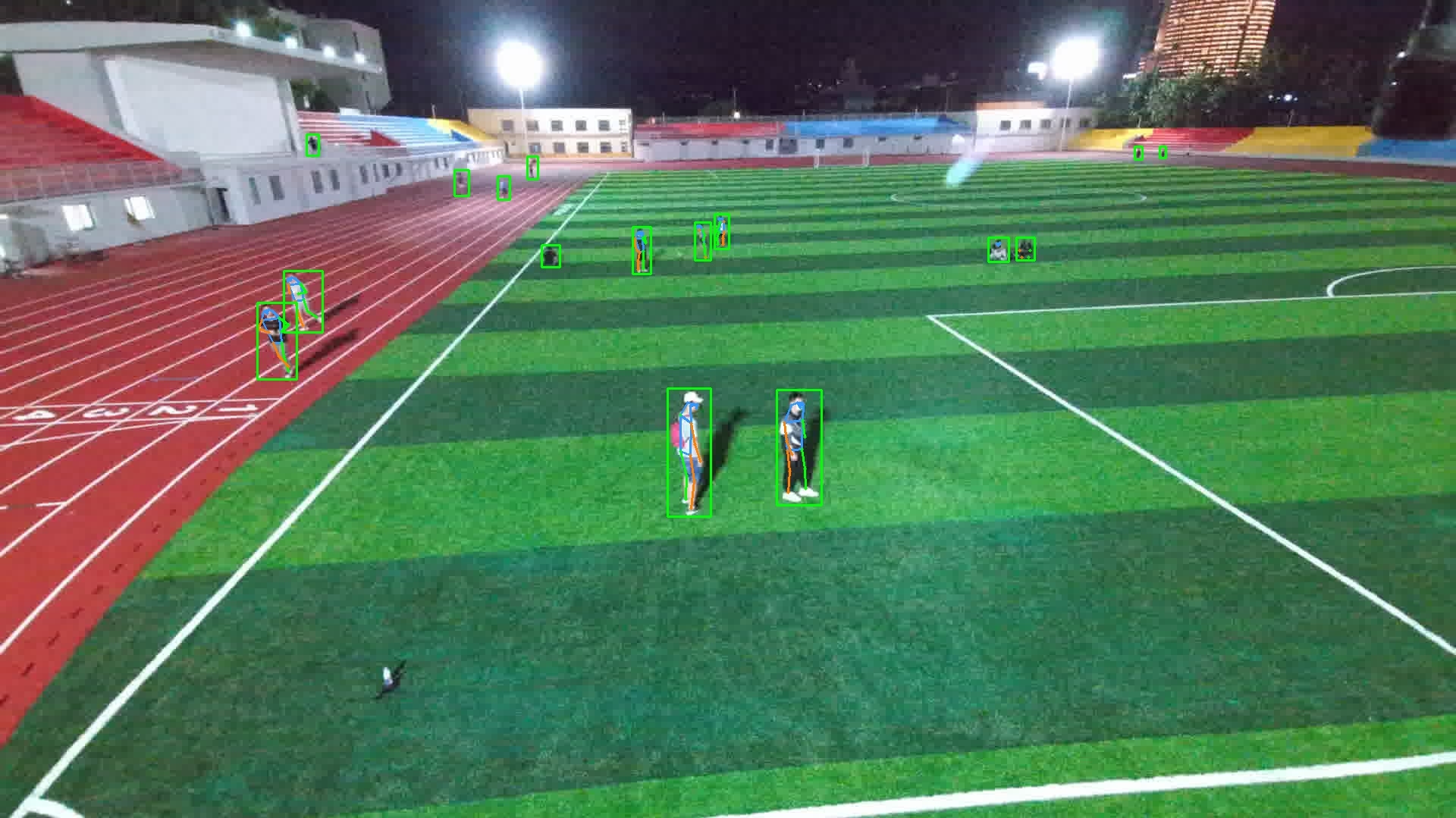}
   \caption{Qualitative FlyPose result on the UAV-Human test-set.}
   \label{fig:result_poseestimation_visdrone}
\end{figure}

ViTPose-S, in particular, achieves the fastest performance on the Jetson Orin with an inference time of just 6.54 ms when converted to a TensorRT FP32 engine, making it the most suitable candidate for our onboard system. Given the realtime demands of various aerial applications, 
ViTPose-S is selected as the core HPE model for \textit{FlyPose}. A qualitative example of the full FlyPose pipeline showing detection and pose estimation outputs is presented in Figure~\ref{fig:result_poseestimation_visdrone}. In combination, our final \textit{FlyPose} model consisting of RT-DETRv2-S and ViTPose-S has an inference latency of only 19.54 milliseconds on a Jetson Orin AGX developer kit for a single image forward pass, excluding pre- or post-processing. 


\subsection{Deployment onboard the UAV}

\begin{figure}[tbp]
  \centering
  \includegraphics[width=\linewidth]{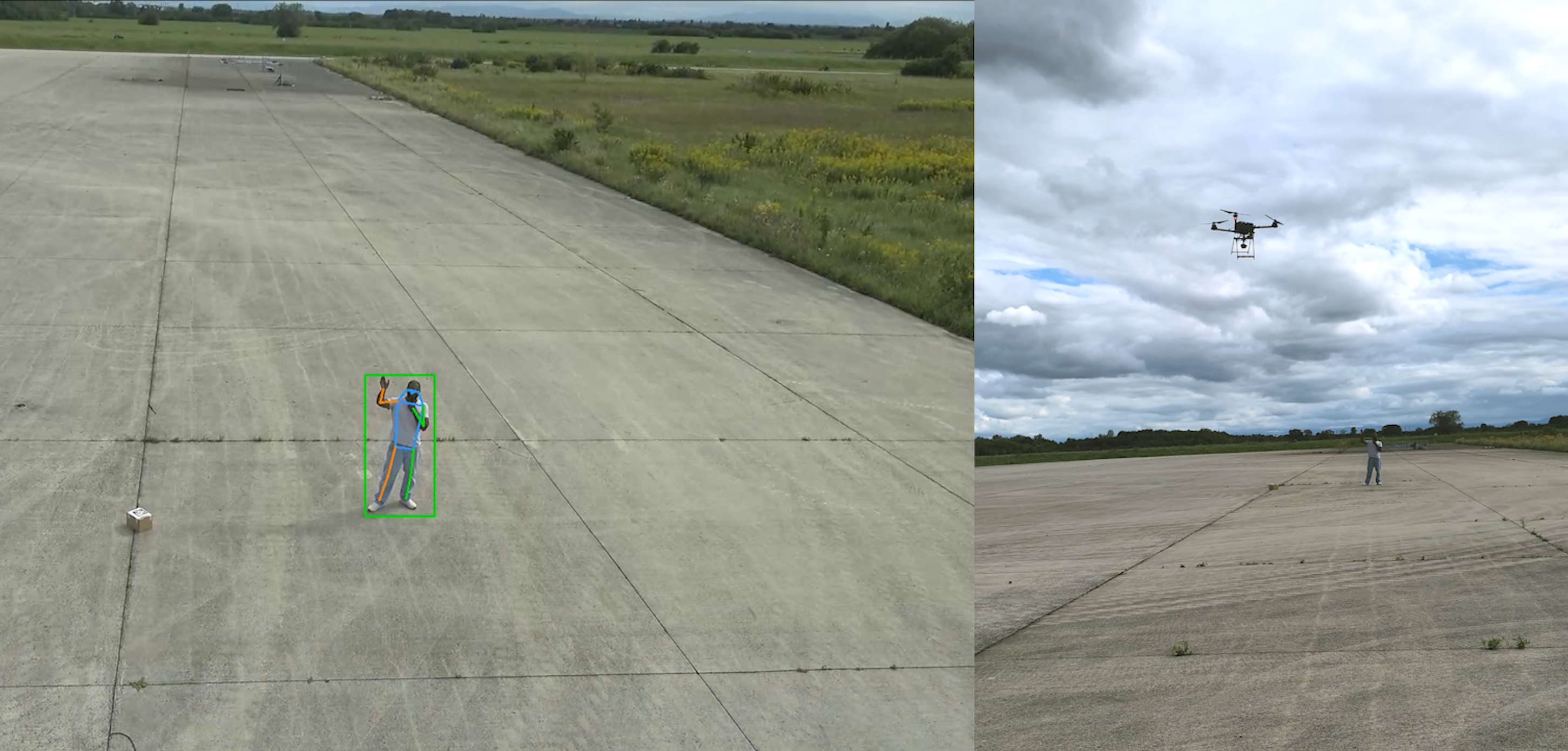}
   \caption{We fly our setup in real flight to demonstrate the feasibility of FlyPose. The synchronized snapshot shows the onboard camera view (left) and the UAV in flight (right). The person detection and predicted poses using FlyPose are overlayed.}
   \label{fig:result_application}
\end{figure}

To validate FlyPose in real flight with a UAV, the Jetson Orin AGX developer kit with 32GB of RAM and a multi-spectral gimbal camera were mounted onto a commercial quadrotor UAV with a maximum take-off weight of 35kg, and connected via Ethernet to receive images from the camera's Full-HD RTSP stream. We then ran FlyPose on the embedded system to predict poses in order to estimate where a person is pointing in a simulated cargo pickup task. Figure \ref{fig:result_application} shows a snapshot from one of the flight experiments, where the left image is a view of the onboard camera and the right shows the UAV itself from an external perspective. 

During our experiments, we observed a one-time initialization delay of approximately 300 ms when acquiring the RTSP image stream from the camera. This latency means that frames become available for processing about 300 ms after the corresponding event occurs. This camera-dependent delay appears only at the start of the stream and can be reduced by using simpler camera interfaces. Once a frame is acquired, the image preprocessing operations (padding and rescaling) performed with the CUDA jetson-utils library take about 0.5 ms per frame. The FlyPose inference requires about 19.54 ms (13 ms for detection, 6.54 ms for pose estimation) for a single person in the scene. We anticipate longer inference times in multi-person scenarios but further experimentation is needed to evaluate the trade-offs of batching detections for pose estimation. With a total time of 20 ms from frame acquisition to pose prediction, the system is well within the range of 25fps realtime performance, leaving roughly another 20 ms per frame for downstream tasks such as tracking, gesture recognition and action recognition.
This is beneficial for novel drone applications that rely on gesture-based human-drone interaction or fast responses to recognized human activity, since these scenarios require low-latency reactions from the UAV.
Our hardware setup remains flexible for additional application-specific hardware as the UAV is capable of carrying 15kg of payload besides its own weight and the flight batteries. The integrated sensory payload currently weighs about 4kg, which leaves 11kg for task-specific hardware.

\section{Discussion}
\label{sec:discussion}

In this section, we discuss the key insights and limitations gathered through the development, evaluation and deployment of \textit{FlyPose}. Through a range of experiments across different model variants and datasets, we explored the challenges within human pose estimation from aerial perspectives.

In the \textbf{person detection} stage of our pipeline, we learned that while training on aerial imagery significantly improved the model's predictions from overhead views, lightweight detectors like RT-DETRv2 still face challenges with small objects, even after training on datasets like TinyPerson. Similar to human perception, in order to distinguish small scale objects from background noise, temporal motion features are sometimes essential. Mixing top-down aerial views with frontal imagery from datasets like COCO frequently led to the model omitting an extended arm when viewing the person from the top or falsely including a background object that resembles an arm. Especially, when the person is wearing darker clothing, the boundary between the persons shadow and the feet for example can be visually ambiguous. Cluttered backgrounds with unfamiliar elements still frequently led to false positives, whereas increasing the confidence threshold also prevented distant objects from being picked up. Although top-down and frontal views, as well as RGB and thermal imagery showcase different visual features, we chose not to separate them into distinct classes. Instead we train on a merged multi-modal dataset, where the person class still represents the same object category, and an applied model should be able to handle viewpoint and color variations. While our detection precision is not on par with larger scale state-of-the-art models on each individual dataset, RT-DETRv2-S still demonstrates good generalization across multiple test sets, views and modalities at a small memory footprint and is realtime feasibile on an embedded system. 

\begin{figure}[H]
  \centering
  \includegraphics[width=\linewidth]{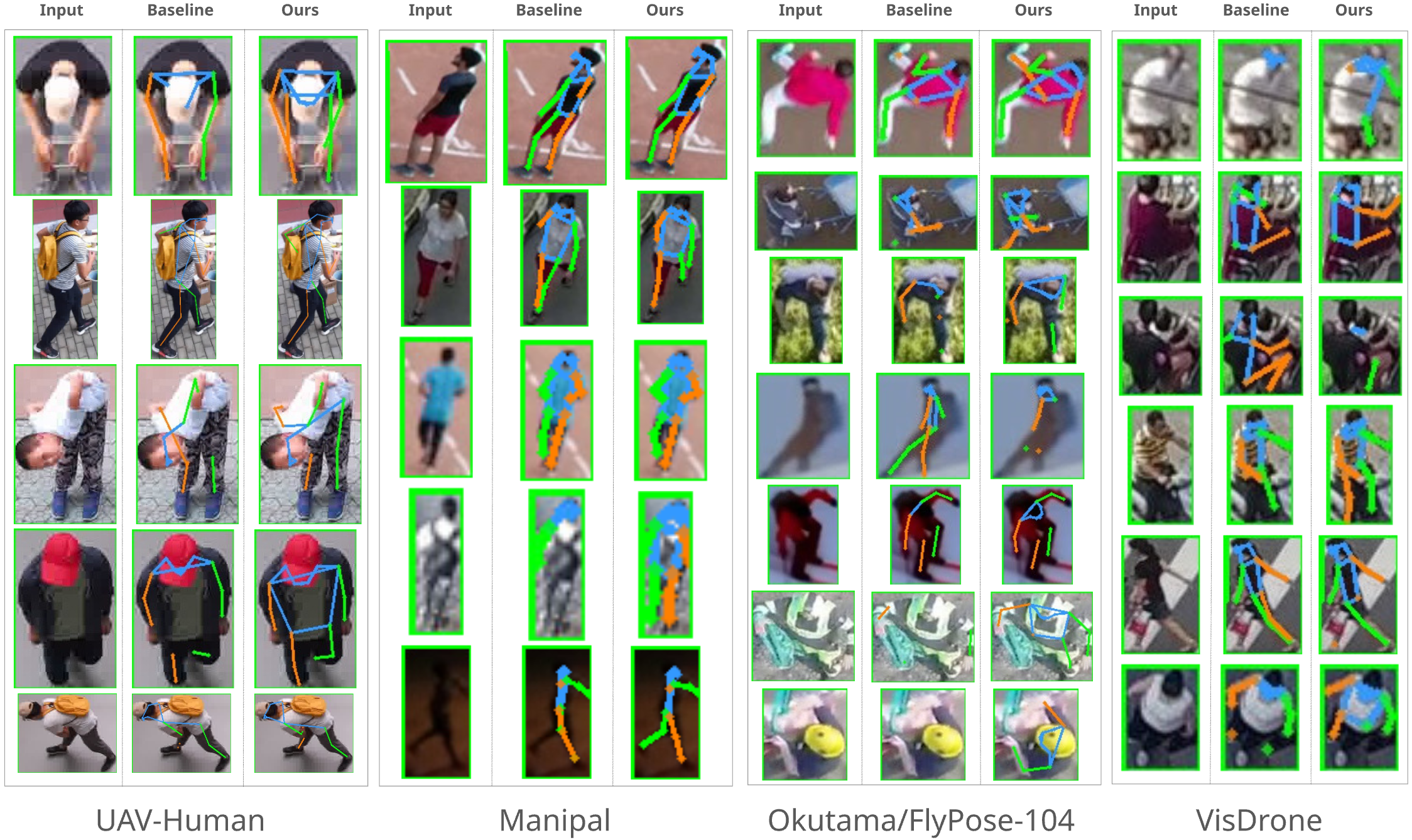}
   \caption{Qualitative FlyPose Results on various aerial datasets. The baseline is the COCO-pretrained ViTPose-S and Ours is the finetuned version.}
   \label{fig:result_qualit_estimation}
\end{figure}

For the second step of \textbf{pose estimation}, we note that HPE from aerial views remains underexplored, primarily due to the scarcity of pose-annotated datasets from this perspective. 
The current state-of-the-art models can greatly improve for aerial perspectives in the presence of good quality datasets. In this context, the FlyPose-104 dataset serves as a particularly challenging benchmark, illustrating how most current models struggle in aerial views.
To address the shortage of aerial pose-annotated datasets, one promising direction could be to leverage transformed 3D HPE datasets to refine predicted poses. While flight altitude and camera metadata are rarely included in existing datasets, these factors could also provide valuable priors to models about expected sizes and resolutions of persons. We additionally attempted to incorporate the PoseFES dataset which offers overhead indoor views into our trainings, but its limited size and constrained diversity did not contribute significantly to the overall model performance. 

\begin{figure*}[htbp]
    \centering
    \begin{subfigure}[t]{0.49\textwidth}
        \centering
        \includegraphics[width=\linewidth]{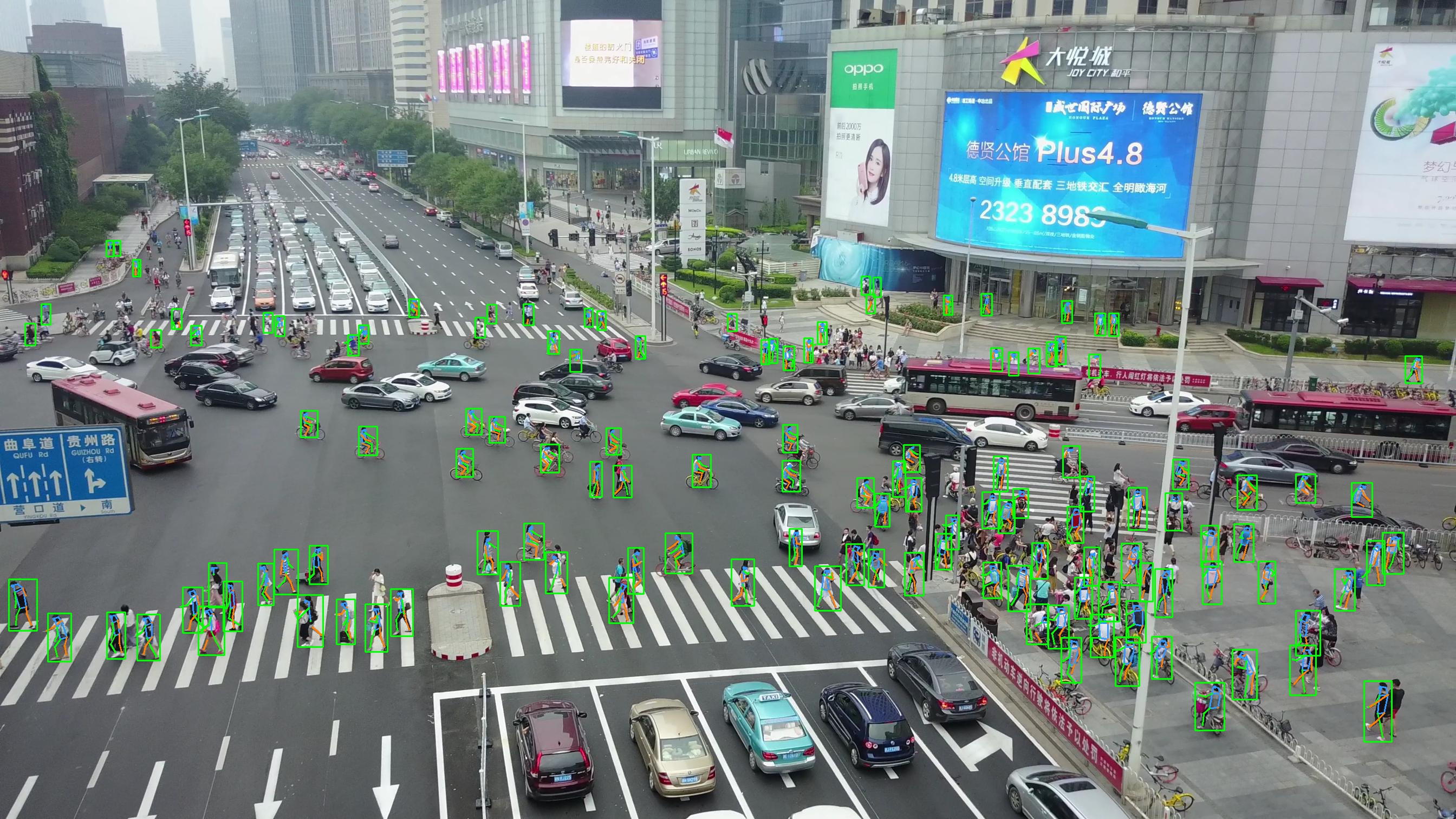}
    \end{subfigure}
    \hfill
    \begin{subfigure}[t]{0.49\textwidth}
        \centering
        \includegraphics[width=\linewidth]{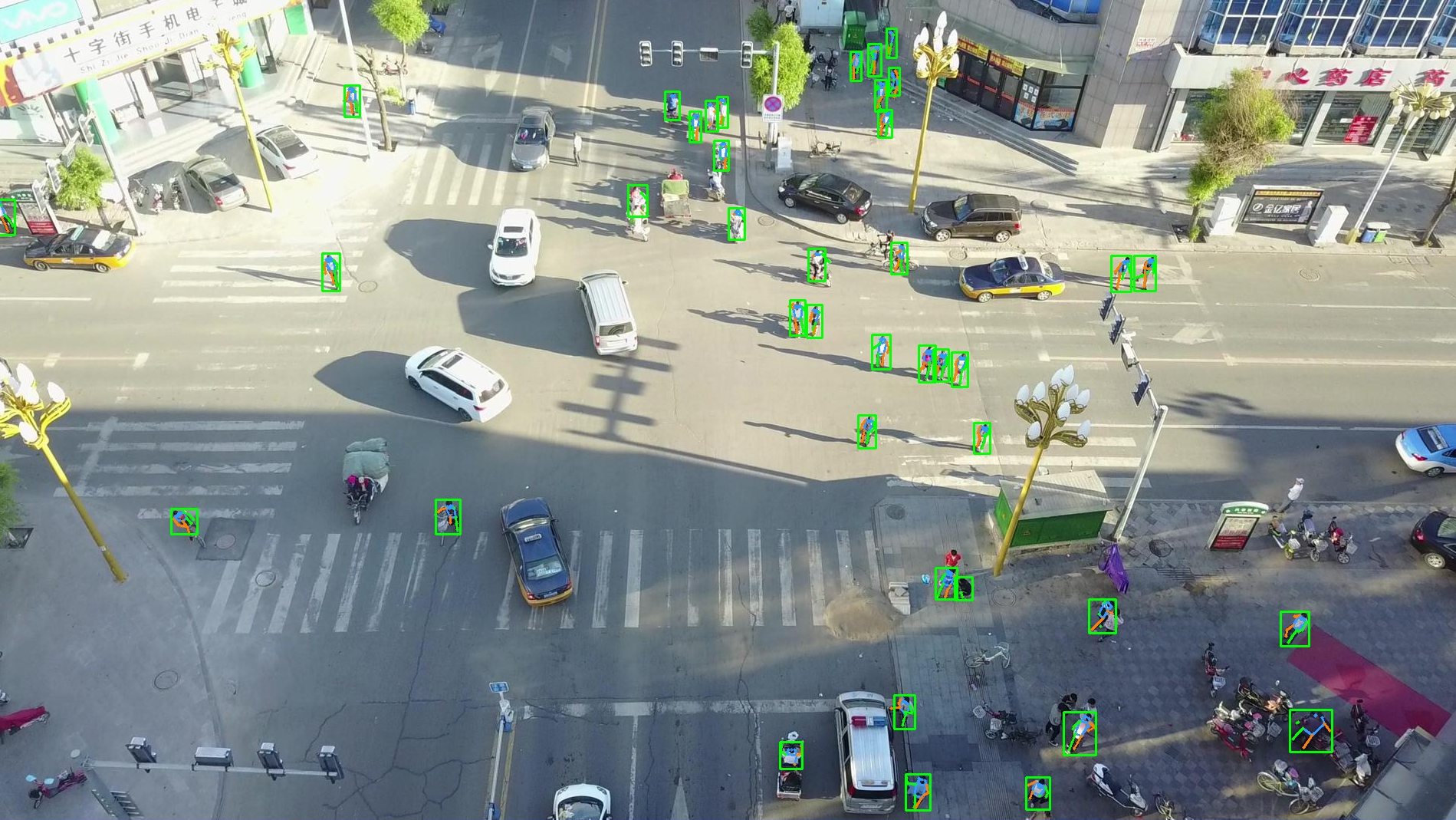}
    \end{subfigure}
    \caption{Qualitative FlyPose results of person detection and pose estimation on the VisDrone dataset.}
    \label{fig:side-by-side}
\end{figure*}

Examples of the pose results of our FlyPose model on the UAV-Human, Manipal, Okutama, FlyPose-104 and VisDrone datasets are shown in Figure \ref{fig:result_qualit_estimation}. The COCO-pretrained ViTPose-S baseline model frequently struggles with discerning self-occluded keypoints, these are better estimated by our finetuned model. It can be observed that it performs relatively better on low resolution patches as well. Cases where our model still struggles are with persons strongly camouflaged by the background or when detected patches contain multiple people in close proximity. Qualitative samples of thermal imagery were also inspected, but while the features did generalize to a degree, the poses were not as reliable as on aerial RGB images. Due to the lack of color information at lower resolutions, left-right keypoint identity swaps occurred more frequently when it was visually ambiguous which direction the person is facing.

We also observe that the keypoint heatmaps show weak confidence for joint detections in low-resolution patches, as they are frequently cut-off by the used confidence threshold of 0.4 for keypoint estimates. Nevertheless, for individuals at greater distances, person detection may suffice for most applications. Our additional evaluation results on the UAV-Human dataset revealed a consistent drop in average OKS scores for facial keypoints such as nose, eyes and ears, compared to body keypoints like shoulders, elbows, wrists, hips, knees and feet. This suggests that facial landmarks are particularly challenging to localize from aerial viewpoints, likely due to occlusions and limited visibility. Although more analysis on the impact of high-person-count images on the system needs to be conducted, one way to deal with this could be an informed selection strategy that only performs keypoint estimation on relevant detected people depending on the current task of the UAV. Finally, despite the current challenges in aerial human pose estimation, FlyPose is a first step to more robustly recognize human poses and perform downstream tasks like gesture and action recognition from aerial views, directly onboard the UAV.


\section{Conclusion}
\label{sec:conclusion}

We presented FlyPose, a lightweight person detection and pose estimation model for aerial views. Our person detector leverages the RT-DETRv2-S architecture, trained on nine aerial detection featuring RGB and thermal imagery. 
It achieves competitive performance across the test sets of VisDrone, Manipal-UAV and HIT-UAV. Our pose estimator utilizes the ViTPose-S architecture and further improves its performance on the UAV-Human dataset.
The FlyPose model delivers quality pose estimation results on a variety of aerial viewpoints, while maintaining a small footprint suitable for embedded systems. The FlyPose model was deployed in real flight experiments to confirm the results from various altitudes and aerial viewpoints.
As drones become increasingly integrated into human-populated environments, the need for intelligent, context-aware drones will grow, so it can be expected that more practically-deployable models like FlyPose tailored for drone applications with realtime requirements will be needed.
\subsection*{Acknowledgements}



This work was funded by the German Federal Ministry for
Economic Affairs and Energy (BMWK), as part of the Federal Aviation Research
Program LuFo VI-2 (CargoPack, Grant-ID: 20D2111D).
{
    \small
    \bibliographystyle{ieeenat_fullname}
    \bibliography{main}
}
\end{document}